\documentclass{article}

\PassOptionsToPackage{numbers, compress}{natbib}

\usepackage[final]{neurips_2019}

\usepackage[utf8]{inputenc} 
\usepackage[T1]{fontenc}    
\usepackage{hyperref}       
\usepackage{url}            
\usepackage{booktabs}       
\usepackage{amsfonts}       
\usepackage{nicefrac}       
\usepackage{microtype}      
\usepackage{algorithm}
\usepackage{algpseudocode}
\usepackage{enumitem}
\usepackage[pdftex]{graphicx}
\usepackage{wrapfig}
\usepackage{makecell}
\usepackage{float}
\setcitestyle{square}

\title{Neural networks on microcontrollers: \\saving memory at inference via operator reordering}

\author{%
  Edgar Liberis$^{1}$\\
  \texttt{edgar.liberis@cs.ox.ac.uk} \\[0.7em]
  $^{1}$Department of Computer Science\\
  University of Oxford, Oxford, UK
  \And
  Nicholas D. Lane$^{1,2}$\\
  \texttt{nicholas.lane@cs.ox.ac.uk} \\[0.7em]
  $^{2}$Samsung AI Center Cambridge\\
  Cambridge, UK
}

\begin{document}

\maketitle

\begin{abstract}
  Designing deep learning models for highly-constrained hardware would allow imbuing many edge devices with intelligence. Microcontrollers (MCUs) are an attractive platform for building smart devices due to their low cost, wide availability, and modest power usage. However, they lack the computational resources to run neural networks as straightforwardly as mobile or server platforms, which necessitates changes to the network architecture and the inference software. In this work, we discuss the deployment and memory concerns of neural networks on MCUs and present a way of saving memory by changing the execution order of the network's operators, which is orthogonal to other compression methods. We publish a tool for reordering operators of TensorFlow Lite models$^{1}$ and demonstrate its utility by sufficiently reducing the memory footprint of a CNN to deploy it on an MCU with 512KB SRAM.
\end{abstract}

\section{Introduction}

Deep learning can bring computational intelligence to personal and IoT devices. Using deep learning models directly on the edge devices allows for greater cost-efficiency, scalability and privacy for end-users, compared to relying on a remote server to carry out the processing. However, the development of lightweight neural networks, suitable for such underpowered hardware, is centred around mobile phones as the target platform.

Here, we venture further and explore a more resource-constrained platform---microcontroller units (MCUs). MCUs are cheap, widespread and are geared towards energy-efficient workloads. They offer an alternative to designing a purpose-built custom chip, allowing to save on development cost and time. However, a unit typically consists of a low-frequency processor and only several hundred kilobytes of on-chip memory~\citep{stm32specs}, and thus severely underpowered compared to mobile devices.

Specially designed network architectures and inference software are required to cope with hardware constraints of MCUs. In this work, we: (a) discuss memory limitations of a microcontroller platform and how it affects neural network deployment; (b) devise a way to minimise the peak memory usage of a neural network by making inference software follow a particular execution order of its operations. 

We implement our methodology as a tool for reordering operators within TensorFlow Lite models.\footnote{Available for download at \url{https://github.com/oxmlsys/tflite-tools}} We successfully apply it to a chosen convolutional neural network, reducing the memory footprint enough to make it fit within the on-chip memory of our microcontroller platform, which would not have been possible using the default provided operator execution order. Operator reordering is carried out only at inference and does not change the architecture or the output of a neural network, making it fully orthogonal to many other network compression methods.

\section{Background}

\subsection{Neural network execution}

A neural network can be thought of as a computation graph which expresses dependencies between individual operations (also called layers or operators). An operator, \emph{e.g.} a 2D convolution or addition, takes one or more input tensors and produces a single output. Modern deep learning frameworks optimise the network’s computation graph for inference in advance by \emph{e.g.} fusing adjacent operators and folding batch normalisation layers into preceding linear operations.
The execution proceeds by evaluating one operator at a time in a topological order of nodes in the graph. 

An operator requires buffers for its inputs and output to be present in memory before its execution can commence. Once the operator has finished executing, memory occupied by its inputs can be reclaimed (if not use elsewhere) and the output buffer will eventually be used as an input to other operators. We define a \emph{working set} as a set of tensors that need to be kept in memory at any given point in execution. This comprises input and output tensors of a pending operator and other tensors that were already produced and need to be held back in memory for subsequent operators.

Classic neural network architectures, such as the original multilayer perceptron, AlexNet~\cite{Krizhevsky2012}, VGG~\cite{VGG}, consist of a linear sequence of layers, which are iteratively applied to transform the input. However, more recent architectures, such as ResNet~\cite{He2016}, Inception~\cite{Szegedy2017}, NasNet~\cite{Zoph2018}, introduce divergent processing paths where the same tensor can be processed by several layers, \emph{i.e.} their computation graph is no longer linear and has branches. This means that the inference software may have multiple operators available for execution at any given step. 

\subsection{Resource scarcity of microcontroller hardware}

A microcontroller unit that would be powerful enough to execute neural networks reasonably quickly (\emph{e.g.} ARM Cortex M series) typically has a low-frequency RISC processing core (up to 400 MHz) and 128--2048KB of on-chip memory~\citep{stm32specs}. The memory is partitioned into read-write static RAM (SRAM) and read-only NOR-Flash memories, with the latter housing the executable code and static data. In contrast to mobile or desktop hardware, there are no intermediate levels in this memory hierarchy, although the set-up may have some backing storage. A backing storage, \emph{e.g.} an SD-card, typically has a high capacity but is slow~\cite{SDBusSpeed} and power-costly to access ($\approx$100x more energy required to read a value outside of on-chip memory~\cite{horowitz20141}). 

The lack of intermediate memories forces applications to fit within the on-chip memory to remain fast. For neural networks, this makes aiming for a small peak working set size and parameter count (model size) an important goal in model design. Note that it’s not necessary to pursue the maximal memory saving---aiming just below the on-chip memory capacity is sufficient.

Memory requirements of a neural network can be directly mapped to the two types of on-chip memory. Parameters (trainable weights, constants) of a network are immutable and can be embedded into the executable code as static data stored in NOR-Flash. Any intermediate tensors that are dependent upon input (so-called activation matrices) are produced at runtime and would have to be stored in SRAM. Thus the model size and peak memory usage are constrained by the capacities of NOR-Flash and SRAM memories, respectively. 

In Section 4, we show how choosing operator execution order affects which tensors reside in SRAM (are in the working set). We exploit this to minimise the peak working set size (peak memory usage).

\section{Related work}

The design of compact models is an active topic of deep learning research, albeit usually under less extreme constraints than those of microcontroller hardware. One can obtain a smaller neural network by using layer decomposition~\cite{lane2016deepx, cai2014fast}, pruning~\cite{Theis2018, deepcompression}, quantisation~\cite{Jacob_2018_CVPR} and binarisation~\cite{alizadeh2018empirical, courbariaux2016binarized}, distillation~\cite{hinton2015distilling} or exploiting sparsity~\cite{Georgiadis2018}. Popular mobile-friendly CNN architectures include MobileNet~\cite{Sandler2018} and ShuffleNet~\cite{Zhang2018}. MNasNet~\cite{Tan2018} and EfficientNet~\cite{Tan2019} develop architecture search algorithms to design a network within a certain floating-point operation count or memory budget. In particular, Fedorov \emph{et al.}~\cite{Fedorov2019} incorporate the maximal working memory size of an operator into their optimisation goal.

A relatively underexplored set of methods include complex evaluation strategies for parts of the network to save memory at runtime. For example, authors of MobileNet~\cite{Sandler2018} note that a building block of their model has a channel-wise operation, whose output is accumulated into another tensor, which allows processing the input tensor in parts. Also, Alwani \emph{et al.}~\cite{alwani2016fused} propose not to materialise an output tensor of a large convolution operation in memory at all, and compute its individual output elements as needed by succeeding operators. 

The development of low-power machine learning models is fostered by the TinyML Summit~\cite{TinyMLSummit} and the Visual Wake Words competition~\cite{Chowdhery2019}, which looked for performant MCU-sized CNNs for person detection. Compact deep learning models have been built for use on wearables devices~\cite{bhattacharya2016sparsification} and for keyword spotting~\cite{zhang2017hello,fernandez2018fly}. Concerns about the memory usage of neural networks and data movement during execution are also discussed in neural network accelerator chip design literature~\cite{Siu2018,Parashar2017,Eyeriss}.

\section{Methods and implementation}

\begin{wrapfigure}[28]{r}{0.32\textwidth}
  \vspace{-2.5em}
  \begin{center}
    \includegraphics[width=0.32\textwidth]{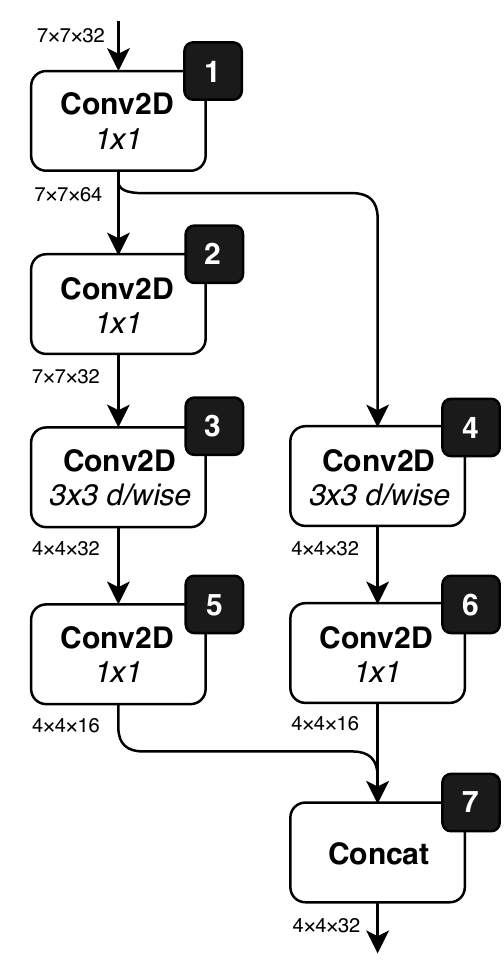}
  \end{center}
\caption{An example computation graph consisting of 1x1 and 3x3 depthwise convolution operators. Annotations on arrows represent tensor sizes.}
\label{fig:computation-graph}
\end{wrapfigure}

Neural networks whose computation graphs contain branches allow some freedom over the order of evaluation of their operators. When execution reaches a branching point, the inference software has to choose which branch to start evaluating next. This choice can affect which tensors need to be kept in memory (working set), so we can construct an execution schedule that minimises the total size of the working set at its peak (memory bottleneck).

To illustrate this, Figure~\ref{fig:computation-graph} shows an example of a computation graph, adapted from a real-world CNN. Evaluating operators as numbered 1 through to 7 will result in peak memory usage of 5216, coming from operator \#3 (input and output buffers + the output of operator \#1 that is held back for operator \#4). However, fully evaluating the rightmost branch first (execution order 1, 4, 6, 2, 3, 5, 7), would result in peak memory usage of 4960, coming from operator \#2 (input and output buffers + output of operator \#6 that is held back for operator \#7). Appendix A gives a more detailed breakdown of the memory usage during computation, together with plots produced by our tool, for both default and optimised operator schedules.

We approach finding a memory-optimal execution schedule for an arbitrary computation graph by algorithmically enumerating all execution schedules and calculating their peak memory usage. To simplify the problem, we assume that no operator will be executed twice (this assumption is also made in TensorFlow).
A computation graph is a directed acyclic graph (DAG) and any topological order of its nodes would produce a valid execution schedule; in general, enumerating all topological orders of a DAG is an explored problem in graph algorithms literature~\cite{knuth1974structured}. 

In Algorithm 1 (procedure \textsc{Mem}), we describe a dynamic programming algorithm that is concerned with the minimal peak memory usage required to produce (and keep in memory) a set of tensors $X$. It enumerates execution schedules recursively by trying to "undo" operators that produced each of the tensors in $X$. The algorithm should be invoked on a set of network’s output tensors and the optimal execution schedule can be traced by checking which recursive calls contributed to the answer. The complexity of the algorithm is $\mathcal{O}(|V|2^{|V|})$, where $|V|$ is the number of operators of the network.

\begin{algorithm}
\newcommand{\LHSComment}[2]{%
\vspace{-1.25em}%
\begin{itemize}[leftmargin=#1]%
\item[$\triangleright$]{#2}%
\end{itemize}%
\vspace{-0.15em}
}%
\caption{Computing the minimal peak memory usage of a neural network. \textsc{Partition} function splits a set into two using a predicate; \textit{producer(x)} denotes an operator that produced tensor $x$.}\label{alg:mem}
\begin{algorithmic}[1]
\Procedure{Mem}{$X$}\Comment{Minimum amount of memory needed to compute tensors in $X$}
  \State \LHSComment{2.5em}{Partition tensors into constants (no producer) and activation matrices}
  \State $cs, as \gets \textsc{Partition}(X, x: \textit{producer}(x) \textbf{ is None})$
  \If{$as $ is empty}
    \State \textbf{return} $\sum_{c \in cs} |c|$ \Comment{No operators left to order, report sizes of remaining constants}
  \EndIf
  \State $m \gets \infty$ \
  \For{$x \textbf{ in } as$} \Comment{Try to unapply the operator that produced $x$}
    \State \textit{rs} $\gets as \setminus x$ \Comment{Remaining tensors that need to be kept in memory}
    \State \textit{is} $\gets \textit{producer}(x)\textit{.inputs}$ \Comment{Tensors required to produce $x$}
    \If{any$(x$ is a predecessor of $r \textbf{ for } r \textbf{ in } rs)$}
      \State \textbf{continue} \Comment{$x$ is a predecessor to $r$, so \textit{producer}$(x)$ would have to be evaluated twice}
    \EndIf
    \State \LHSComment{4em}{Peak memory usage will be determined by either the producer of $x$---\emph{i.e.} memory used its input tensors ($is$), output tensor ($x$) and other tensors ($rs$)---or by other operators in the execution path (recursive case $\textsc{Mem}(rs \cup is)$)}
    \State $m' \gets \max(\textsc{Mem}(rs \cup is), \sum_{t \in rs \cup is \cup \{x\}} |t|)$
    \State $m \gets \min(m, m')$ \Comment{Pick the execution path that gives minimal memory usage}
  \EndFor
  \State \textbf{return} $\sum_{c \in cs} |c| + m$ 
\EndProcedure
\end{algorithmic}
\end{algorithm}

To simplify the implementation, the algorithm begins by filtering out tensors that don’t have an operator that produced them (so-called constants), as those just contribute to memory usage and don’t affect the execution schedule. A restriction that no operator is evaluated twice is implemented by checking whether an operator is a predecessor to any of the remaining tensors, as this would require it to be executed at some point again in the schedule. Note that \textsc{Mem}$(X)$ may be invoked on the same set of tensors multiple times (from different execution paths), so it should be memoized (\emph{i.e.} the output should be cached) to avoid recomputing the result. 

We use a lightweight TensorFlow Light Micro inference engine~\cite{TFLMRepo} (henceforth micro-interpreter) to run the neural network on the MCU itself. At the time of writing\footnote{At the time of publication of this pre-print, a dynamic memory allocator has been implemented by maintainers of TensorFlow Lite Micro, making this change no longer necessary. However, we keep the description of our memory allocation strategy, as well as the power and latency measurements, as an illustrative example of memory management overheads.}, the software did not support reclaiming memory from tensors that were no longer needed, so we implement our own dynamic memory allocator for tensor buffers.
Internally, TensorFlow Lite assumes that tensors reside in contiguous blocks of memory and cannot be fragmented. The memory allocator is only used by the micro-interpreter, which allows us to ensure that C/C++ pointers to memory blocks are not being remembered anywhere in the code. This enables us to move buffers in memory as needed for defragmentation. We adopt a very simple defragmentation strategy of moving all tensor buffers to the start of the memory region as much as possible after the execution of every operator. 

\section{Experiments}
We deploy a neural network onto an MCU with both default and optimised operator schedules to exhibit the difference in memory usage. We use one of the winning submissions of the Visual Wake Words competition~\cite{Chowdhery2019}, called SwiftNet Cell~\cite{ShiftNetVVW,Cheng2019}, as it has only 250KB of parameters and contains many branches, which enables us to showcase the benefits of reordering. The model is run using the modified micro-interpreter (as described above) on a NUCLEO-F767ZI prototyping board~\cite{NucleoBoard}. The board features a Cortex M7 processor, running at 216Mhz, and has 512KB of SRAM.

\begin{table}[H]
    \centering
    \renewcommand{\arraystretch}{1.1}
    \begin{tabular}{lrrrr}
    \toprule
    & \multicolumn{2}{c}{SwiftNet Cell} & \multicolumn{2}{c}{MobileNet v1} \\
    & Default order
    & Optimal order
    & Static alloc.
    & Dynamic alloc. \\
    \cmidrule(lr){2-3} \cmidrule(lr){4-5}
\makecell[l]{Peak memory usage\\(excl. overheads)} & 
351KB &
301KB &
241KB &
55KB ($\downarrow$ 186KB) \\
Execution time &
\textit{N/A} &
10243 ms &
1316 ms &
1325 ms ($\uparrow$ 0.68\%)\\
Energy use &
\textit{N/A} &
8775 mJ &
728 mJ &
735 mJ ($\uparrow$ 0.97\%) \\ 
\bottomrule
\end{tabular}
\vspace{2mm}
\renewcommand{\arraystretch}{1.0}
\caption{Peak memory usage, execution time and energy use of chosen models.}
\vspace{-5mm}
\label{tab:results}
\end{table}

Table~\ref{tab:results} shows that optimised ordering was able to save 50KB of memory, compared to the order embedded in the model. Including the framework overhead ($\approx$200KB for SwiftNet Cell, proportional to the number of tensors), this made a sufficient difference to make the model’s memory footprint fit within SRAM.\footnote{The authors of the model note that peak memory usage can be lowered, likely by using fused operator implementations. Further savings would come from optimising the memory usage of the micro-interpreter itself.} We also check the overhead introduced by replacing a static memory allocator with a dynamic one by running MobileNet-v1-based~\cite{Howard2017} person detection model (from the Tensorflow Lite Micro repository~\cite{TFLMRepo}). Measurements show negligible (sub-1\%) increase in execution time and energy used by the MCU and the memory footprint was decreased by 186KB. In general, latency and power consumption can be reduced with operator implementations that leverage processor capabilities well (SIMD, DSP instructions).

\section{Discussion}
The results show that employing a different operator execution order for neural network inference can make previously undeployable models fit within the memory constraints of MCU hardware. Reordering operators can be implemented just within the inference software, making it orthogonal to most other network compression approaches, which were likely to have been already used to create an MCU-sized model.   

Unlike mobile and server platforms, MCU hardware often doesn’t have enough memory to statically pre-allocate all tensor buffers of the network, which requires the inference software to support dynamic memory allocation. We showed that a simple defragmentation strategy is a viable option with little overhead cost. However, when the execution schedule is known in advance, optimal tensor buffer placement in memory may be precomputed. 

Having a way of precisely computing peak memory usage for models with complex computation graphs would benefit neural architecture search (NAS) procedures. The algorithm can be extended to support various memory saving tricks: for example, if one of the inputs to the addition operator is not used elsewhere, the result can be accumulated into it, eliminating the need for an output buffer.

\section{Conclusion}
Microcontrollers are a viable platform for running deep learning applications if the model designer can overcome constraints imposed by limited memory and storage. In this work, we describe how to minimise peak memory usage of a neural network during inference by changing the evaluation order of its operators. By applying our methodology, we were able to achieve sufficient memory savings to deploy the chosen CNN on a microcontroller with 512KB SRAM, which would not have been possible otherwise. Our tool for embedding optimal operator ordering into TensorFlow Lite models is available at \url{https://github.com/oxmlsys/tflite-tools}.

\subsubsection*{Acknowledgments}

The work was supported by the Engineering and Physical Sciences Research Council UK (EPSRC UK), grant ref. EP/S001530/1, and Samsung AI. The authors would also like to thank Javier Fernández-Marqués for providing help with measuring energy usage.


\medskip

\small

\bibliography{operator-reordering.bib}
\bibliographystyle{ieeetr}

\newpage
\section*{Appendix A}

\begin{figure}[H]
\begin{minipage}{0.5\linewidth}
    \centering
    \includegraphics[width=\linewidth]{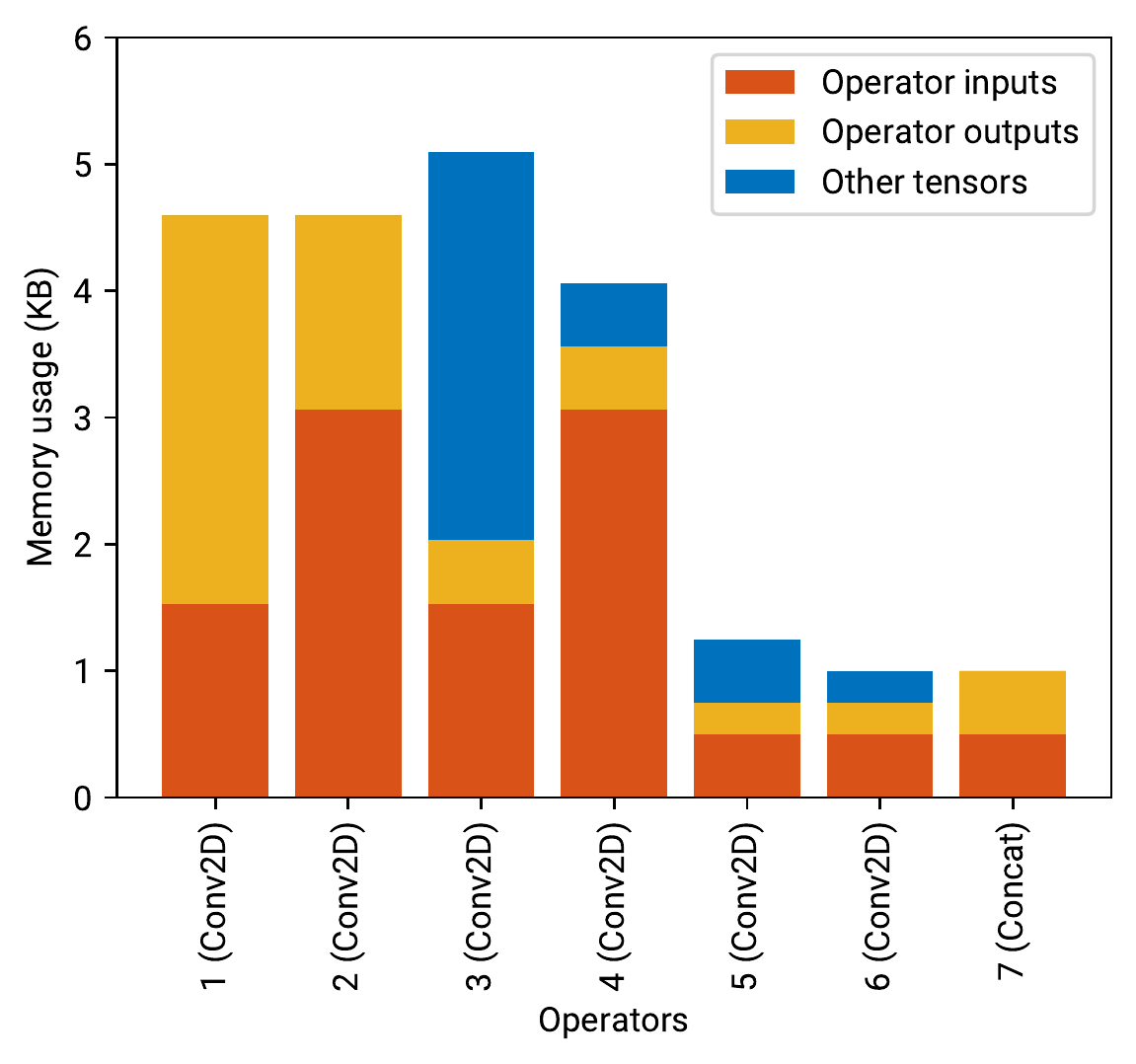}
\end{minipage}
\begin{minipage}{0.5\linewidth}
    \centering
    \vspace{-2.8em}
    \begin{tabular}{ccr}
    \toprule
    Operator & \makecell{Tensors in RAM \\ (output of op. \#)} & Usage (B) \\
    \midrule
    1 (\texttt{Conv2D}) & $\{0, 1\}$    & 4,704 \\
    2 (\texttt{Conv2D}) & $\{1, 2\}$    & 4,704 \\
    3 (\texttt{Conv2D}) & $\{1, 2, 3\}$ & 5,216 \\
    4 (\texttt{Conv2D}) & $\{1, 3, 4\}$ & 4,160 \\
    5 (\texttt{Conv2D}) & $\{3, 4, 5\}$ & 1,280 \\
    6 (\texttt{Conv2D}) & $\{4, 5, 6\}$ & 1,024 \\
    7 (\texttt{Concat}) & $\{5, 6, 7\}$ & 1,024 \\
    \midrule
    & \makecell[r]{\textbf{Peak:}} & 5,216 \\
    \bottomrule
    \end{tabular}
    \\\vspace{0.8em}
    Memory usage w/ default operator order
\end{minipage}

\caption{Memory usage of the sample computation graph with default operator ordering.}
\label{fig:default_order}
\end{figure}

\begin{figure}[H]
\begin{minipage}{0.5\linewidth}
    \centering
    \includegraphics[width=\linewidth]{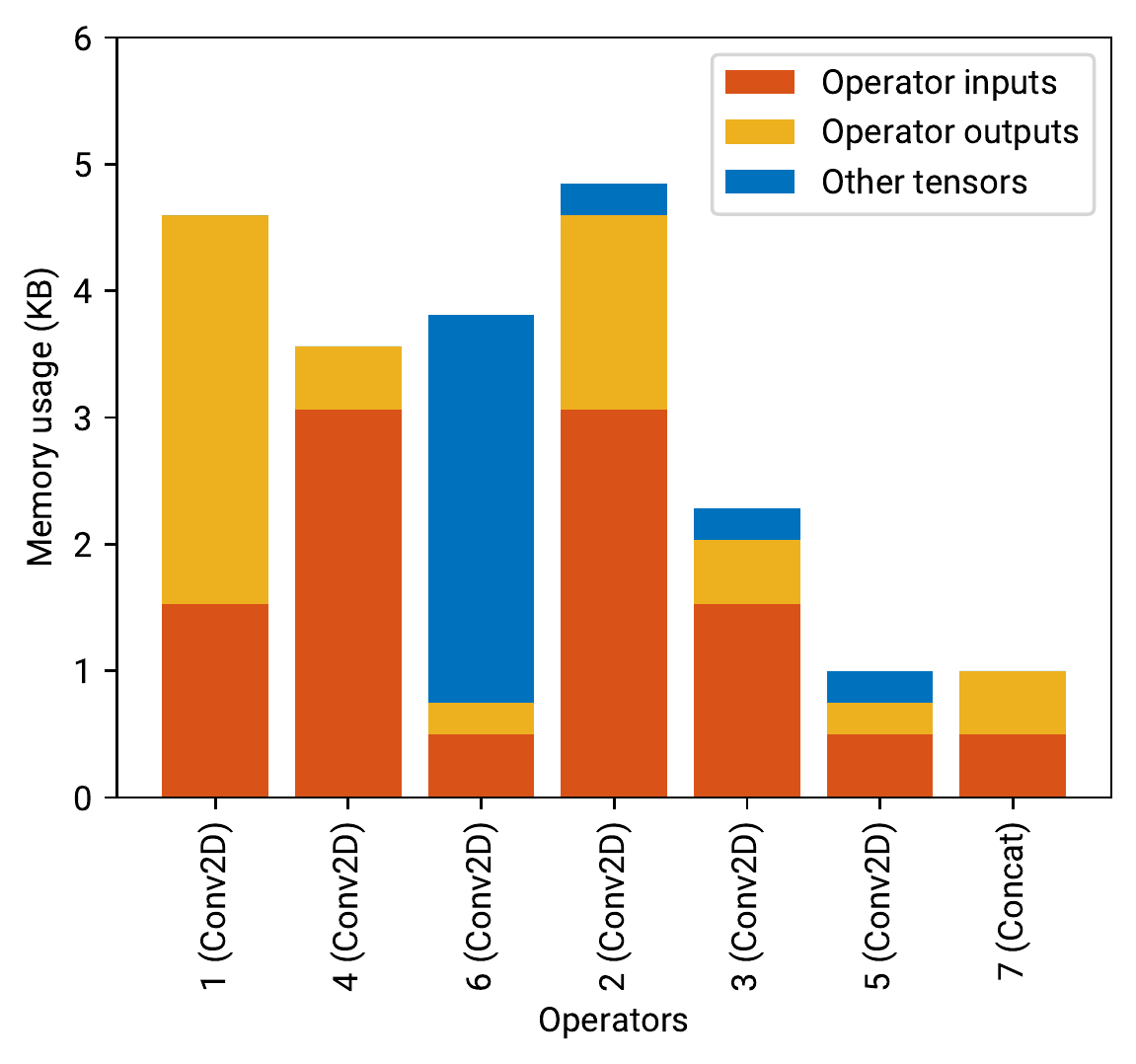}
\end{minipage}
\begin{minipage}{0.5\linewidth}
    \centering
    \vspace{-2.8em}
    \begin{tabular}{ccr}
    \toprule
    Operator & \makecell{Tensors in RAM \\ (output of op. \#)} & Usage (B) \\
    \midrule
    1 (\texttt{Conv2D}) & $\{0, 1\}$    & 4,704 \\
    4 (\texttt{Conv2D}) & $\{1, 4\}$    & 3,648 \\
    6 (\texttt{Conv2D}) & $\{1, 4, 6\}$ & 3,904 \\
    2 (\texttt{Conv2D}) & $\{1, 2, 6\}$ & 4,960 \\
    3 (\texttt{Conv2D}) & $\{2, 3, 6\}$ & 2,336 \\
    5 (\texttt{Conv2D}) & $\{3, 5, 6\}$ & 1,024 \\
    7 (\texttt{Concat}) & $\{5, 6, 7\}$ & 1,024 \\
    \midrule
    & \makecell[r]{\textbf{Peak:}} & 4,960 \\
    \bottomrule
    \end{tabular}
    \\\vspace{0.8em}
    Memory usage w/ optimised operator order
\end{minipage}

\caption{Memory usage of the sample computation graph with optimised operator ordering.}
\label{fig:optimized_order}
\end{figure}

\end{document}